\documentclass[conference]{IEEEtran}

\usepackage{amsmath}
\usepackage{lipsum}
\usepackage{multicol}
\usepackage[pdftex]{graphicx}
\usepackage{graphicx}
\usepackage[export]{adjustbox}

\usepackage{float}
\floatstyle{plaintop}
\restylefloat{table}
\usepackage{placeins}
\usepackage{graphicx}
\usepackage{wrapfig}
\usepackage[labelfont=bf]{caption}
\usepackage{color}
\usepackage{tabulary}
\usepackage{makecell}
\usepackage{array} 
\usepackage{amsmath}
\usepackage{amsthm}
\usepackage{algpseudocode}
\usepackage{algorithm}
\usepackage{wrapfig}
\usepackage{array, tabularx, caption, boldline}
\usepackage{graphicx}
\usepackage{cellspace}

\usepackage[font={footnotesize}]{caption}
\usepackage{setspace}

\ifCLASSINFOpdf
\else
\fi

\begin{document}
	
	\title{\textsc{\small  Accepted Manuscript in IEEE International Conference on Machine Learning and Applications (ICMLA 2017)}\\
		SUBIC: A Supervised Bi-Clustering Approach for Precision Medicine}
	
	\author{\IEEEauthorblockN{Milad Zafar Nezhad$^{a}$, Dongxiao Zhu$^{b,*}$, Najibesadat Sadati$^{a}$, Kai Yang$^{a}$, Phillip Levy$^{c}$}\\
		\IEEEauthorblockA{Department of Industrial and Systems Engineering,
			Wayne State University$^{a}$\\Department of Computer Science, Wayne State University$^{b}$\\Department of Emergency Medicine and Cardiovascular Research Institute, Medical School, Wayne State University$^{c}$\\
			Corresponding author$^{*}$, E-mail addresses: dzhu@wayne.edu}
		\and
	}
		
	\maketitle

\begin{abstract}
		
Traditional medicine typically applies one-size-fits-all treatment for the entire patient population whereas precision medicine develops tailored treatment schemes for different patient subgroups. The fact that some factors may be more significant for a specific patient subgroup motivates clinicians and medical researchers to develop new approaches to subgroup detection and analysis, which is an effective strategy to personalize treatment. In this study, we propose a novel patient subgroup detection method, called Supervised Biclustring (SUBIC) using convex optimization and apply our approach to detect patient subgroups and prioritize risk factors for hypertension (HTN) in a vulnerable demographic subgroup (African-American). Our approach not only finds patient subgroups with guidance of a clinically relevant target variable but also identifies and prioritizes risk factors by pursuing sparsity of the input variables and encouraging similarity among the input variables and between the input and target variables.\\

		
\end{abstract}

\IEEEpeerreviewmaketitle
\section{Introduction}
The explosive increase of Electronic Medical Records (EMR) and emerge of precision (personalized) medicine in recent years holds a great promise for greatly improving quality of healthcare \cite{bochicchio2016big}. In fact, the paradigm in medicine and healthcare is transferring from disease-centered (empirical) to patient-centered, the latter is called Personalized Medicine. The extensive and rich patient-centered data enables data scientists and medical researchers to carry out their research in the field of personalized medicine \cite{nezhad2016safs}. Personalized medicine is defined as \cite{redekop2013faces}: ``use of combined knowledge (genetic or otherwise) about an individual to predict disease susceptibility, disease prognosis, or treatment response and thereby improve that individual health." In other words, the goal of personalized medicine is to provide the right treatment policy to the right patient at the right time. 

A crucial step in personalized medicine is to discover the most important input variables (disease risk factors) related to each patient \cite{deng2014budgeted}. Since identification of risk factors needs multi-disciplinary knowledge including data science tools, statistics techniques and medical knowledge, many machine learning and data mining methods have been proposed to identify, select and prioritize risk factors \cite{aguwa2017modeling}\cite{sadati2017observational}\cite{roostaei2017spatially}. Some popular methods such as linear model with shrinkage \cite{tibshirani1996regression} and random forest \cite{breiman2001random} effectively select significant risk factors for the entire patient population. However, these approaches are not capable of detecting risk factors for each patient subgroup because they are developed based on an assumption that the patient population is homogeneous with a common set of risk factors. 

While the point of input variable selection is well taken, the association with small subgroups, a key notion in personalized medicine, is often neglected. As mentioned, personalized healthcare aims to identify subgroup of patients who are similar with each other according to both target variables and input variables. Discovering potential subgroups plays a significant role in designing personalized treatment schemes for each subgroup. Therefore, it is essential to develop a core systematic approach for patient subgroup detection based on both input and target variables \cite{fan2016new}. A number of data-driven approaches have been developed for subgroup identification. The more popular methods can be divided in two categories: 1) Tree-based approaches \cite{doove2014comparison} (or so called recursive partitioning), and 2) Biclustering approaches \cite{pontes2015biclustering}. Tree based methods in subgroup analysis are greatly developed in recent years, such as Model-based recursive partitioning \cite{zeileis2008model}, Interaction Trees \cite{su2008interaction}, Simultaneous Threshold Interaction Modeling Algorithm (STIMA) \cite{dusseldorp2010combining}, Subgroup Identification based on Differential Effect Search (SIDES) \cite{lipkovich2011subgroup}, Virtual Twins \cite{foster2011subgroup}, Qualitative Interaction Tree (QUINT) \cite{dusseldorp2014qualitative} and Subgroup Detection Tree \cite{li2017sdt}. The second approaches (Biclustering) have been extensively developed and applied to analyze gene expression data. Most of the biclustering algorithms developed up-to-date are based on optimization procedures as the search heuristics to find the subgroup of genes or patients.

Tree-based methods detect patient subgroups using the relationship between input and target variables whereas biclustering methods just focus on clustering rows and columns of the input variables simultaneously to identify different subgroups with specific risk factors (prioritized input variables). The former employs a target variable to guide subgroup detection by selecting a common set of input variables. The latter selects subgroup of specific input variables without guidance of a target variable. Moreover, both approaches are heuristic in nature that subgroup detection and risk factor identification are sensitive to choices of data sets and initializations hence has a poor generalization performance. Our proposed method combines the strength of the both approaches by using a target variable to guide the subgroup detection and selecting subgroup of specific risk factors. Meanwhile, our systematic approach overcomes the stability limitation of both approaches by casting the problem into a stable and mature convex optimization framework. Figure \ref{figure0} demonstrates consecutive steps of our approach: 
\begin{figure}[H]
	\centering
	\graphicspath{ {figure1} }
	\captionsetup{justification=centering,margin=2.5cm}
	\includegraphics[scale= 0.4]{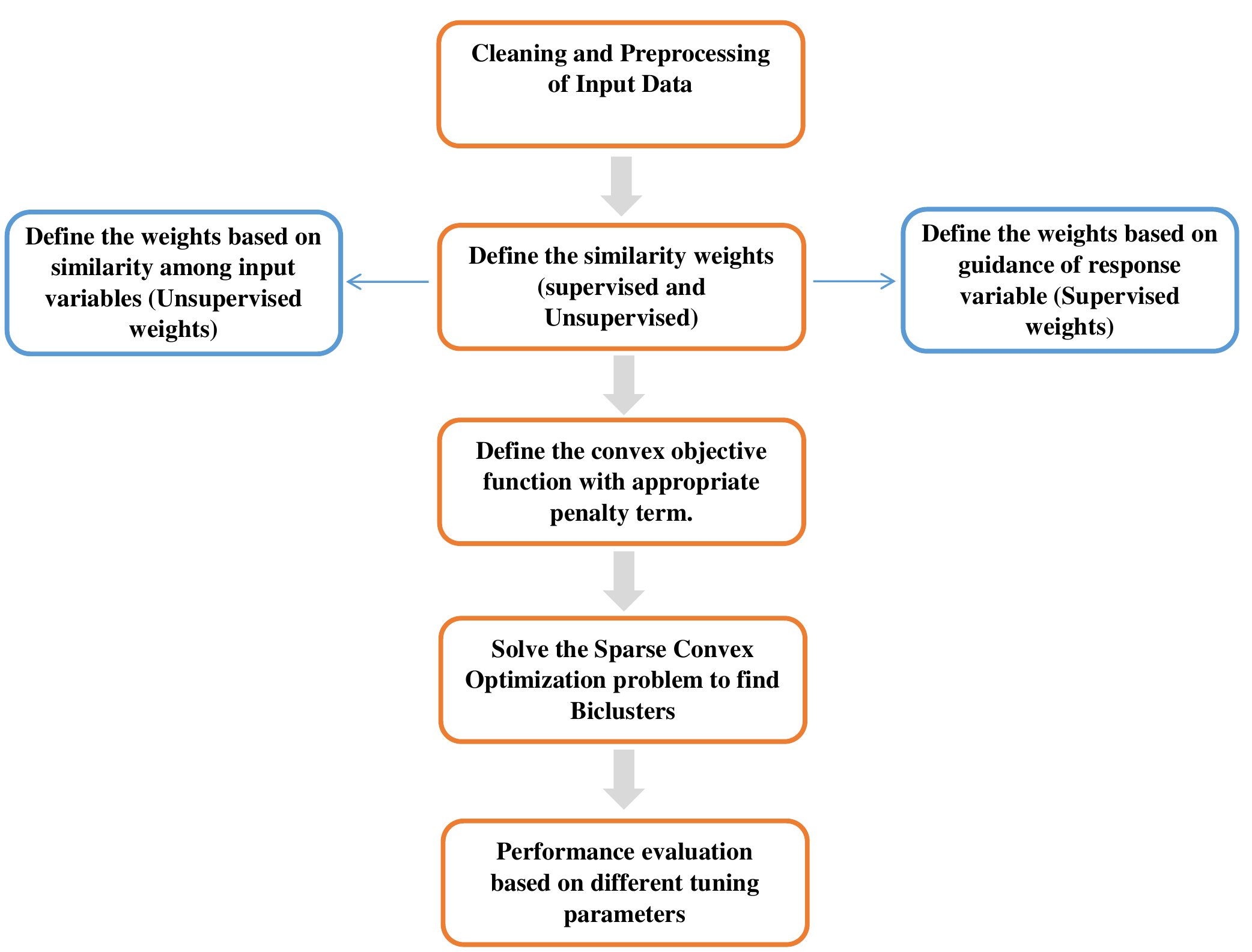} 
	\caption{The consecutive steps of our approach}
	\label{figure0}
\end{figure}
In this study, we propose a new supervised biclustering approach, called SUBIC, for solving patient subgroup detection problem. Our approach is the generalized (supervised) version of convex biclustering \cite{chi2016convex}, which enables prediction of target variables for new input variables. Moreover, we employ the elastic-net penalty \cite{zou2005regularization} (both $l_{1}$ and $l_{2}$ regularization terms) that encourages sparsity of the correlated input variable groups ($X$) with the guidance of a target value ($Y$). Our model is specifically designed for patient subgroup detection and target variable prediction from high dimension data. To the best of our knowledge, our model is the first supervised biclustering approach that can be applied in many domains such as personalized medicine. To demonstrate the performance of SUBIC approach, we apply it to detect subgroups among hypertension (HTN) patients with guidance of left ventricular mass indexed to body surface area (LVMI), a clinically important target variable.  

The rest of this paper is organized as follows. Section II reviews the related works in unsupervised biclustering approaches. Section III explains our proposed supervised biclustering (SUBIC) approach. Section IV describes experimental studies and model evaluation using simulation studies. Section V reports the results of application of our method on patient at the high risk of cardiovascular disease, and finally we conclude this study in Section VI. 

\section{Related Works}
Biclustering is defined as simultaneous clustering of both rows and columns in the input data matrix. Such clusters are important since they not only discover the correlated rows, but also identify the group of rows that do not behave similarly in all columns \cite{eren2013comparative}. In the context of precision medicine, rows correspond to patients and columns correspond to input variables measured in each patient. Biclustering was originally introduced in 1972 \cite{hartigan1972direct}, and Cheng and Church \cite{cheng2000biclustering} were the first to develop a biclustering algorithm and applied it to gene expression data analysis. There exist a wide range of biclustering methods developed using different mathematical and algorithmic approaches. Tanay et al. \cite{tanay2002discovering} proved that biclustering is a NP-hard problem, and much more complicated than clustering problem \cite{divina2006biclustering}. Therefore, most of methods are developed based on heuristic optimization procedures \cite{pontes2015biclustering}. Madeira and Oliveira \cite{madeira2004biclustering}, Busygin et al. \cite{busygin2008biclustering}, Eren et al. \cite{eren2013comparative} and Pontes et al. \cite{pontes2015biclustering} provided four comprehensive reviews about biclustering methods in 2004, 2008, 2012 and 2015 respectively. Based on the most recent review \cite{pontes2015biclustering}, biclustering approaches can be divided in two main groups. The first one refers to methods based on evaluation measures, which means some heuristic methods are developed using a measure of quality to reduce the solution space and complexity of biclustering problem. Table \ref{table1} demonstrates different algorithmic categories within this group:

\begin{table}[H]
	\scriptsize 
	\centering
	\caption{Biclustering methods based on evaluation measure.}\label{table1}
	\begin{tabular}{p{1.2 cm}|p{3.1cm}|p{3.2cm}}
		\hline
		\textbf{Algorithm}    & \textbf{Description}  & \textbf{Prosperous Methods} \\ \hline
		Iterative greedy search  &These methods follow a greedy strategy to find an approximate solution. They improve the measure of evaluation in each step and construct a set of objects from the smallest possible solution space recursively or iteratively. & Direct Clustering \cite{hartigan1972direct}, Cheng and Church \cite{cheng2000biclustering}, HARP Algorithm \cite{yip2004harp}, Maximum Similarity Bicluster\cite{liu2007computing}  \\ \hline
		
		Stochastic iterative greedy search  & These methods use a stochastic strategy by adding a random variable to the iterative greedy search in order to speed up the biclustering algorithm.  & Flexible Overlapped Biclustering \cite{yang2005improved}, Random Walk Biclustering \cite{angiulli2008random}, Reactive GRASP Biclustering \cite{dharan2009biclustering}, Pattern-Driven Neighborhood Search\cite{ayadi2012pattern} \\ \hline
		
		Nature-inspired meta-heuristics  &These methods are developed based on a nature-inspired meta-heuristic, such as simulated annealing, ants colony and swarm optimization. & Simulated-Annealing Biclustering \cite{bryan2006application}, Evolutionary Algorithms for Biclustering \cite{bleuler2004ea}, SEBI(Sequential Evolutionary Biclustering) \cite{divina2006biclustering}, Multi-objective-Evolutionary Algorithms for Biclustering\cite{mitra2006multi}, Bicluster Ensemble using Mutual Information \cite{aggarwal2013bemi} \\ \hline
		
		Clustering-based approach  & These methods carry out their search based on traditional clustering methods in one dimension and then use an additional approach to cluster second dimension.& Possibilistic-Spectral Biclustering. \cite{cano2007possibilistic}, Biclustering with SVD and Hierarchical Clustering.\cite{yang2011finding} \\ \hline
		
	\end{tabular}
\end{table}

The second group of approaches is called non metric-based biclustering methods that do not use any measure of quality (evaluation measure) for guiding the search. These methods use graph-based or probabilistic algorithms to identify the patterns of biclusters in data matrix. Table \ref{table2} summarizes different algorithms of non metric-based group: \\

\begin{table}[H]
	\scriptsize  
	\centering
	\label{table2}
	\caption{Biclustering methods based on non-metric.}\label{table2}
	\begin{tabular}{p{1.2 cm}|p{3.1cm}|p{3.2cm}}
		\hline
		\textbf{Algorithm}    & \textbf{Description}  & \textbf{Prosperous Methods} \\ \hline
		Graph-based approaches &These methods are developed based on the graph theory. They use nodes for either genes, samples or both gene and sample representations, or refer to nodes as representing the whole biclusters.   & Statistical-Algorithmic Method for Bicluster Analysis (SAMBA)\cite{tanay2002discovering}, Qualitative-Biclustering algorithm (QUBIC)\cite{li2009qubic}, Pattern-based Co-Regulated Biclustering (CoBi) \cite{roy2013cobi}, MicroCluster \cite{zhao2005microcluster}\\ \hline
		
		One-way clustering-based approaches   &These methods are developed based on the same concept of clustering-based approached, but they do not use any measure of quality in their search path.   & Coupled Two-way Clustering \cite{getz2000coupled}, Interrelated Two-way Clustering \cite{tang2005interrelated} \\ \hline
		
		Probabilistic search &These methods are created using statistical modeling and probability theory. & Plaid Models [58], Rich Probabilistic Models \cite{segal2001rich}, Gibbs Sampling \cite{sheng2003biclustering}, Bayesian Biclustering Model \cite{gu2008bayesian} \\ \hline
		
		Linear algebra  & These methods use linear algebra to apply linear mapping between vector spaces for describing and identifying the most correlated submatrices from the original dataset.& Spectral Biclustering \cite{kluger2003spectral}, Iterative Signature Algorithm \cite{bergmann2003iterative}, Non-smooth Non-negative Matrix Factorization (nsNMF) \cite{carmona2006biclustering} \\ \hline
		
		Optimal reordering rows and columns   & These methods are based on the strategy of performing permutations of the original rows and columns in the data matrix, to achieve a better arrangement and make biclusters.& Pattern-based Biclustering \cite{henriques2014bicpam}, Order-Preserving-Sub-Matrices (OPSMs)\cite{ben2003discovering}\\ \hline
		
	\end{tabular}
\end{table}

One of the important aspects of bicluster structure is overlapping, which means several biclusters share rows and columns with each other. Because of the characteristic of search strategy in biclustering methods, overlapping may or may not be allowed among the biclusters. Most of the algorithms mentioned in Table \ref{table1} and Table \ref{table2} allow overlapping biclusters \cite{pontes2015biclustering}. Since these algorithms use heuristic approach for guiding search, final biclusters may vary depending on how the algorithm is initialized. Therefore, they don't guarantee a global optimum nor are they robust against even small perturbations \cite{chi2016convex}. 

Recently, Chi et al. \cite{chi2016convex} formulated biclustering problem as a convex optimization problem and solved it with an iterative algorithm. Their convex biclustering model corresponds to checkerboard mean model, which means each data matrix component is assigned to one bicluster. They used the concept of fused lasso \cite{tibshirani2005sparsity} and generalized it with a new sparsity penalty term corresponding to the problem of convex biclustering. This method has some important advantages over the previous heuristic-based methods, that is, it created a unique global minimizer for biclustering problem, which maps data to one biclustering structure, therefore the solution is stable and unique. Also it used a single tuning parameter to control the number of biclusters. Authors performed simulation studies to compare their algorithm with two other biclustering algorithms, dynamic tree cutting algorithm \cite{langfelder2008defining} and sparse biclustering algorithm \cite{tan2014sparse}, which assume the checkerboard mean structure. Results showed that convex biclustering outperforms the competing approaches in terms of Rand Index \cite{chi2016convex}. 

Despite the improved performance, the convex biclustering method, like other biclustering methods, does not exploit a target variable on subgroup detection and risk factor selection. As a result, the detected biclusters do not link to target variables of interest. Hence, it is unable to predict the target variable for future input variables. Clearly, the target variable such as LVMI provides a critical guidance for detection and selection of the meaningful biclusters (patient subgroups). 
Moreover, the $l_{1}$ penalty term alone in convex biclustering encourages the sparsity of individual input variables but overlooks the fact that they are also correlated within variable groups. To overcome both limitations, we introduce a new elastic-net regularization term that seeks sparsity of the correlated variable groups and employs a target variable to supervise the biclustering optimization process. Consequently, our model is truly a predictive model that is capable of predicting value of the target variable for new patients. In the next section, we describe our method in detail.

\section{Method}
\subsection {The object function of the SUBIC method}

Let's assume that the input data matrix $X_{n \times p}$ represents $n$ instances with different $p$ input variables and $Y_{n}$ is the continuous target variable (e.g. LVMI), corresponds to $n^{th}$ instance (patients). According to the checkerboard mean structure, we assume $R$ and $C$ are the sets of rows and columns of the bicluster $B$ respectively, and $x_{i,j}$ refers to elements belong to the bicluster $B$, the observed value of $x_{i,j}$ can be defined as \cite{chi2016convex}: $x_{i,j} = \mu_{0} + \mu_{RC} + \varepsilon_{i,j} $, where $\mu_{0}$ is a baseline mean for all elements, $\mu_{RC}$ is the mean of bicluster corresponds to R and C, and $\varepsilon_{i,j} $ refers to error that is i.i.d with $N(0, \sigma)$. With considering non-overlapping biclusters, this structure corresponds to a checkerboard mean model \cite{kluger2003spectral}. Without loss of generality, we ignore $\mu_{0}$ from all elements. The goal of biclustering is to find the partition indices with regard to $R$ and $C$ then estimate the mean of each corresponding bicluster ($B$). To achieve this goal, we minimize the following convex objective function: 
\begin{align}\label{eq:general1}
&\text{\footnotesize $F_{\lambda_{1},\lambda_{2}} = \frac{1}{2} \| X-T \|_{F}^{2} + P(T)$},&
\end{align}
where matrix $T \in R^{n\times p}$ includes our optimization parameters, which are the estimate of means matrix. The first term is frobenius norm of matrix $X-T$ refers to error term and $P(T) = P_1{(T)}+ P_2{(T)} $ is the elastic-net regularization penalty term formulated as follows:
\begin{align}
& \text{\footnotesize $P_1{(T)}= \lambda _{1} [\Sigma _ {i<j} w_{i,j} \|T_{.i} - T_{.j}\|_ {2}^{2} +\Sigma _ {i<j} h_{i,j} \|T_{i.} - T_{j.}\|_ {2}^{2} ]$},&
\end{align}
and
\begin{align}
& \text{\footnotesize $P_2{(T)}= \lambda _{2} [\Sigma _ {i<j} w_{i,j} \|T_{.i} - T_{.j}\|_ {1} +\Sigma _ {i<j} h_{i,j} \|T_{i.} - T_{j.}\|_ {1} ]$}.&
\end{align}
It is clear that this objective function is similar to subset selection problem in regularized regression \cite{tibshirani1996regression}. In the penalty function $\lambda _{1}$ and $\lambda _{2}$ are tunning parameters. The first term penalized by $\lambda _{1}$ is a $l_{2}$-norm regularization term and the second term  penalized by $\lambda _{2}$ is a $l_{1}$-norm regularization term. Therefore the penalty term $P(T)$ acts as regression elastic-net penalty \cite{zou2005regularization}. $T_{i.}$ and $T_{.i}$ refer to $i$th row and column of matrix $T$, which can be considered as a cluster center (centroid) of $i$th row and column respectively. 
By minimizing the objective function defined in Eq.\ref{eq:general1} with sparsity based regularization, the cluster centroids are shrunk together when the tunning parameters increase. It means that sparse optimization tries to unify the similar rows and columns to specific centroid simultaneously. Finding the similarity between rows and columns is guided by different weights ($w_{i,j}$, $h_{i,j}$), which are included in objective function. These weights has been defined based on distance between input variables ($X_{.i} - X_{.j}$ and $X_{i.} - X_{j.}$), distance between target variables ($Y_{i} - Y_{j}$) and correlation between input variables and target variable ($X_{.i} , Y_{.j}$). Therefore both input variables and target variable play significant rule in guiding of sparsity to find the best centroids. The first kind of weights ($w_{i,j}$) proceeds the columns convergence and the second one ($h_{i,j}$) proceeds the rows convergence. The weights are constructed from un-supervised and supervised parts, where:
\begin{align}
& \text{\footnotesize$ w_{i,j} = w_{i,j}^{1} + w_{i,j}^{2} \quad \text{and} \quad h_{i,j} = h_{i,j}^{1} + h_{i,j}^{2}$}.&
\end{align}
The unsupervised part ($w_{i,j}^{1}, h_{i,j}^{1}$) attempts to converge rows (columns) based on the similarity exists among input variables, and the supervised part ($w_{i,j}^{2}, h_{i,j}^{2}$) converges rows and columns according to the similarity of input and target variables. Since the rows and columns are in $R^{n}$ and $R^{p}$ spaces respectively, it is required to normalize the weights (recommended the sum of row weights and column weights to be $\frac{1}{\sqrt{n}}$ and $\frac{1}{\sqrt{p}}$ respectively). We used the idea of sparse Gaussian kernel weights \cite{chi2016convex} for defining $w_{i,j}^{1}, w_{i,j}^{2}, h_{i,j}^{1}, h_{i,j}^{2}$. Table \ref{table3} demonstrates the mathematical description of weights:

\begin{table}[H]
	\scriptsize 
	\centering
	\caption{Description of the weights formula.}\label{table3}
	\begin{tabular}{c|p{6.5cm}}
		\hline
		\textbf{$\#$} & \textbf{Weight Formula} \\ \hline
		1 & $w_{i,j}^{1} = l_{i,j}^{k} \exp ^{(-\varphi \|X_{.i} - X_{.j}\|_ {2}^{2})}$ \newline \newline ${*}$This weight is to converge the similar columns in terms of distance similarity measure.  $l_{i,j}^{k}$ is 1 when $j^{th}$ column is among the $k$-nearest neighbor of $i^{th}$ column, otherwise it is zero. Therefore it guarantees the weights are sparse. ($0\leq \varphi \leq1$)   \\ \hline
		
		2  &$w_{i,j}^{2} = l_{i,j}^{k} \exp ^{(-\varphi |corr(x_{.i}, Y) - corr(x_{.j},Y)|)}$ \newline \newline ${*}$This weight is the supervised part of $w_{i,j}$, the goal is to converge the columns that have similar correlation with target variable. It means that the features which behave similarly with target variable should be converged. In our model we used Pearson correlation that assumes a linear relationship between input variable and target variables. ($0\leq \varphi \leq1$)  \\ \hline
		
		3 &$h_{i,j}^{1} = l_{i,j}^{k} \exp ^{(-\varphi \|X_{i.} - X_{j.}\|_ {2}^{2})}$ \newline \newline ${*}$This weight is the same as $w_{i,j}^{1}$, which attempts to converge the similar rows with lower distance from each other.($0\leq \varphi \leq1$) \\ \hline
		
		4  & $h_{i,j}^{2} = l_{i,j}^{k} \exp ^{(-\varphi \sqrt{|(Y_{i} -Y_{j)}}|)}$ \newline \newline ${*}$This weight is supervised part of $h_{i,j}$, and it converges the rows that are similar in term of target variable value. This weight considers the role of target variable in clustering of similar rows.($0\leq \varphi \leq1$) \\ \hline

	\end{tabular}
\end{table}

The way to define the weights has a substantial impact on the quality of biclustering. The weights described above guarantee the sparsity of the problem and employ the similarity of all input and target variables in supervised and unsupervised manner. According to defined weights, the two columns (rows) that are more similar with each other will get larger weight in the convex penalty function, therefore in minimization process, those columns (rows) should be in higher priority, and it means that convex minimizer attempts to cluster the similar columns (rows). The choice of elastic-net penalty term can overcome the lasso limitations. While the $l_{1}$-norm can generates a sparse model, the quadratic part of the penalty term encourages grouping effect and stabilizes the $l_{1}$-norm regularization path. Also the elastic-net regularization term would be very suitable for high dimension data with correlated input variables \cite{zou2005regularization}.

\subsection{The algorithm to train the SUBIC model}
It can be proved easily that the objective function in Eq.\ref{eq:general1} is a convex function. Therefore we need to develop appropriate algorithm to solve this unconstrained convex optimization. Since the second part of penalty function, $P_{2}(T)$ is undifferentiated we use Split Bregman method \cite{ye2011split} developed for large-scale Fused Lasso. It can be shown that this method is equivalent to the alternating direction method of multipliers (ADMM) \cite{boyd2011distributed}. Readers can refer to Split Bregman method \cite{ye2011split} or ADMM algorithm \cite{boyd2011distributed} for more comprehensive explanation. According to both methods we need to use splitting variable and Lagrangian multiplier and then apply augmented Lagrangian for undifferentiated part ($P_{2}(T)$) of objective function. First we need to transform our problem to the equality-constrained convex optimization problem by defining two new variables ($V$,$S$) and adding two constraints correspond to $P_{2}(T)$ and then use Lagrangian multipliers:
\begin{flalign} \label{eq:min}
& \text{\footnotesize $\mathrm{min} \quad F_{\lambda_{1},\lambda_{2}} = \frac{1}{2} \| X-T \|_{F}^{2} + \lambda _{1} [\Sigma _ {i<j} w_{i,j} \|T_{.i} - T_{.j}\|_ {2}^{2} +$} & \notag\\ 
& \text{\footnotesize $\qquad \qquad \quad \: \: \: \: \:\Sigma _ {i<j} h_{i,j} \|T_{i.} - T_{j.}\|_ {2}^{2} ] +\lambda _{2} [\Sigma _ {i<j} w_{i,j} \|T_{.i} - T_{.j}\|_ {1}+$} & \notag\\ 
& \text{\footnotesize $ \qquad \qquad \quad \: \: \: \: \:\Sigma _ {i<j} h_{i,j} \|T_{i.} - T_{j.}\|_ {1} ] $},& \notag\\ 
&\text{\footnotesize $ \text{subject\:to}: \: \: w_{i,j} (T_{.i} - T_{.j}) = V_{i,j} \quad \forall i, j; i<j $}, &\notag\\
&\text{\footnotesize $\qquad \qquad \quad \: \: h_{i,j} (T_{i.} - T_{j.}) = S_{i,j}\quad \forall i, j; i<j $},&
\end{flalign}
where $V$ and $S$ are matrices in $R^{n\times p}$. Assuming the differentiated part of objective function in (1) is $F^{'}_{\lambda_{1},\lambda_{2}}$, the Lagrangian Multiplier for the above problem is:
\begin{flalign}
& \text{\footnotesize $\tilde{L} (T, M, N, V, S) =F^{'}_{\lambda_{1},\lambda_{2}} + \lambda _{2} [\Sigma _ {i<j} w_{i,j} \|V_{i,j}\|_ {1} +\Sigma_{i<j} h_{i,j} \|S_{i,j}\|_ {1} ]+$}&\notag\\
& \text{\footnotesize $\Sigma _ {i<j}\left\langle M_{i,j} ,w_{i,j} (T_{.i} - T_{.j}) - V_{i,j}  \right\rangle +\Sigma _ {i<j}\left\langle N_{i,j} ,h_{i,j} (T_{i.} - T_{j.}) - S_{i,j}  \right\rangle$},&
\end{flalign}
where $M$ and $N$ are the vectors of dual variables (Lagrangian Multipliers) corresponding with each constraints in Eq.\ref{eq:min} (totally there are ${n \choose 2} + {p \choose 2}$ constraints). Finally the Augmented Lagrangian function of Eq.\ref{eq:min}  is as following:
\begin{flalign}
& \text{\footnotesize ${L} (T, M, N, V, S) =F^{'}_{\lambda_{1},\lambda_{2}} + \lambda _{2} [\Sigma _ {i<j} w_{i,j} \|V_{i,j}\|_ {1} +\Sigma _ {i<j} h_{i,j} \|S_{i,j}\|_ {1} ] +$}& \notag\\
& \text{\footnotesize $\Sigma _ {i<j}\left\langle M_{i,j} ,w_{i,j} (T_{.i} - T_{.j}) - V_{i,j}  \right\rangle + \Sigma _ {i<j}\left\langle N_{i,j} ,h_{i,j} (T_{i.} - T_{j.}) - S_{i,j} \right\rangle+$}&\notag\\ 
& \text{\footnotesize $\frac{\mu_{1}}{2} [\Sigma _ {i<j} \|w_{i,j} (T_{.i} - T_{.j}) - V_{i,j}\|_ {2}^{2}] + \frac{\mu_{2}}{2} [\Sigma _ {i<j} \|h_{i,j} (T_{i.} - T_{j.}) - S_{i,j}\|_ {2}^{2}]$},&
\end{flalign}
where $\mu_{1}>0$ and $\mu_{2}>0$ are two parameters. The Split Bregman algorithm for supervised convex biclustering problem described below:

\begin{algorithm}[H]
	\scriptsize  
	\caption{\small Split Bregman algorithm for Training the SUBIC model}
	\begin{algorithmic}[1]
		\State  Initialize $T^{0}$, $V^{0}$,$S^{0}$, $M^{0}$, $N^{0}$
		\Repeat
		\State 
		$\begin{aligned}[t]
		&T^{k+1} =\mathrm{argmin}_{T}\quad \frac{1}{2} \| X-T \|_{F}^{2} + \lambda _{1} [\Sigma _ {i<j} w_{i,j} \|T_{.i} - T_{.j}\|_ {2}^{2}+  & \\
		&\Sigma _ {i<j} h_{i,j} \|T_{i.} - T_{j.}\|_ {2}^{2} ]+ \Sigma _ {i<j}\left\langle M_{i,j}^{k} ,w_{i,j} (T_{.i} - T_{.j}) - V_{i,j}^{k} \right\rangle + & \\
		& \Sigma _ {i<j}\left\langle N_{i,j}^{k} ,h_{i,j} (T_{i.} - T_{j.}) - S_{i,j}^{k}  \right\rangle + \frac{\mu_{1}}{2} [\Sigma _ {i<j} \|w_{i,j} (T_{.i} - T_{.j}) &\\ 
		& - V_{i,j}^{k}\|_ {2}^{2}] + \frac{\mu_{2}}{2} [\Sigma _ {i<j} \|h_{i,j} (T_{i.} - T_{j.}) - S_{i,j}^{k}\|_ {2}^{2}]&     
		\end{aligned} $ \newline \newline

		\State $V_{i,j}^{k+1}= \tau_{\frac{\lambda_{2}}{\mu_{1}}} (w_{i,j} (T_{.i}^{k+1} - T_{.j}^{k+1})+ \mu_{1}^{-1} M_{i,j}^{k}) \quad i<j $ \newline
		
		\State $S_{i,j}^{k+1}= \tau_{\frac{\lambda_{2}}{\mu_{2}}} (h_{i,j} (T_{i.}^{k+1} - T_{j.}^{k+1})+ \mu_{2}^{-1} N_{i,j}^{k}) \quad i<j $ \newline
		
		\State $M_{i,j}^{k+1}= M_{i,j}^{k} + \delta_{1} (w_{i,j} (T_{.i}^{k+1} - T_{.j}^{k+1}) - V_{i,j}^{k+1}) \quad i<j; 0<\delta_{1} \leq \mu_{1} $ \newline
		
		\State $N_{i,j}^{k+1}= N_{i,j}^{k} + \delta_{2} (h_{i,j} (T_{i.}^{k+1} - T_{j.}^{k+1}) - S_{i,j}^{k+1}) \quad i<j;  0<\delta_{2} \leq \mu_{2} $ \newline
		
		\Until \\ Convergence
		
	\end{algorithmic}
\end{algorithm}

$\tau$ acts as a soft thresholding operator defined on vector space and satisfying the following equation:
\begin{flalign}
&\tau_{\lambda}(w)= [t_{\lambda}(w_{1}), t_{\lambda}(w_{2}), ...]^{T},&\notag\\
& \text{where:} \quad t_{\lambda}(w_{i}) = \mathrm{sgn}(w_{i}) \mathrm{max} \{0,|w{i}-\lambda|\}.&
\end{flalign}

\subsection{The SUBIC based prediction approach}
For prediction of the target variable based on supervised biclustering framework, we introduce a simple yet effective approach based generalized additive model (GAM) \cite{hastie1990generalized}. Assuming that $K$ biclusters 
\{$BC_{1}$,$BC_{2}$,..., $BC_{K}$ \} are detected by training the SUBIC model, we consider $K$ classifiers corresponding to each biclusters, i.e., $f_{k} (y|x_{bc_{k}}, x_{\text{new}}) = \overline{y}_{bc_{k}} $. It means that each classifier predicts the target value as an average of the target variables of the corresponding bicluster. The proposed GAM model is as follows:
\begin{align}\label{eq:general}
&g(E(y)) = R_{1}(x_{bc_{1}})+ R_{2}(x_{bc_{2}}) +... +  R_{k}(x_{bc_{k}}),\:& \notag\\ 
&\text{where}\quad R_{k}(x_{bc_{k}}) = q_{k} f_{k}(y|x_{bc}, x_{\text{new}}).&
\end{align}
$q_{k}$ is defined as normalized weight based on posterior probabilities. Assuming that each bicluster follows a Gaussian distribution as $N(\mu_{i}, \sigma)$ and $P(bc_{k}|x_{new})$ is the posterior probability which refers to the probability of each bicluster given a new instance $𝑥$, we can define $q_{k}$ as below:
\begin{flalign}\label{eq:general}
&\text{$q_{k}=\frac{P(bc_{k}|x_{\text{new}})}{\sum _{i=1}^{k} P(bc_{i}|x_{\text{new}})},$}&\notag\\
&\text{where:}\quad P(bc_{k}|x_{\text{new}}) = P(x_{\text{new}}|bc_{k}) \times P(bc_{k}). & 
\end{flalign}
$P(x_{\text{new}}|bc_{k})$ is conveniently calculated based on Gaussian distribution assuming equal variance and zero covariance and $P(bc_{k})$ is the prior that can be calculated by counting the number of instances in each bicluster. 

\section{Experimental Study and Model Evaluation }
For assessing the performance of our approach, we carry out simulation studies and use Rand Index (RI) \cite{rand1971objective} and Adjusted rand index (ARI) \cite{hubert1985comparing} as two popular measures for evaluating the quality of clustering. Since our biclustering method is supervised, 
we simulate data for input and target variables based on a checkerboard mean structure. We used normal distribution with different means to generate simulated data. Figure \ref{fig:checkboard} illustrates an  example simulation study.

As shown below, data was simulated in $20\times20$ matrix. Data in each segment has different size and were created based on a different normal distribution, all sections are generated with low-noisy data ($\sigma=1.5$). Input data in segments (2, 3, 4 and 5) are in high positive correlation with the target variable and input data in segment (6, 7, 8 and 9) are in high negative correlation with the target variable. Segments 1 and 10 in general, are similar with very low correlation with target variable. Segments 1 and 3 of the target variable are positive and the other two sections have negative values. 

According to this assumptions and consider the effect of target variable, it is clear that the true number of biclusters should be 16 (not 10). It means that segments 1 and 10 include 4 biclusters within each. The results of SUBIC implementation for different tuning parameters are displayed in Figure \ref{fig:sim}.
\begin{figure}[H]
	\centering
	\includegraphics[scale= 0.45]{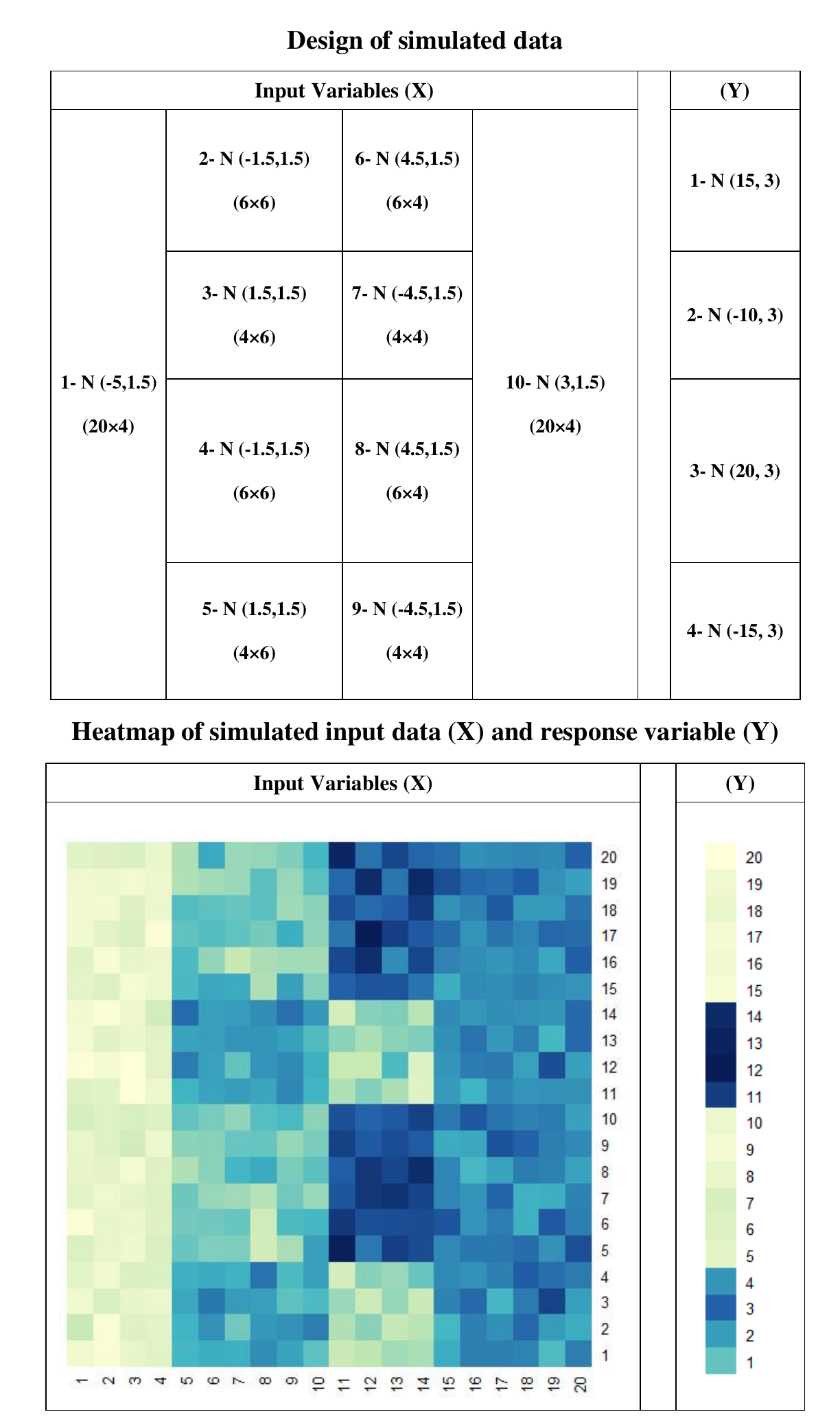} 
	\caption{The chessboard structure and the simulated data (top and down panel).}
	\label{fig:checkboard}
\end{figure}

\begin{figure*}
	\centering
	\includegraphics[scale= 0.43]{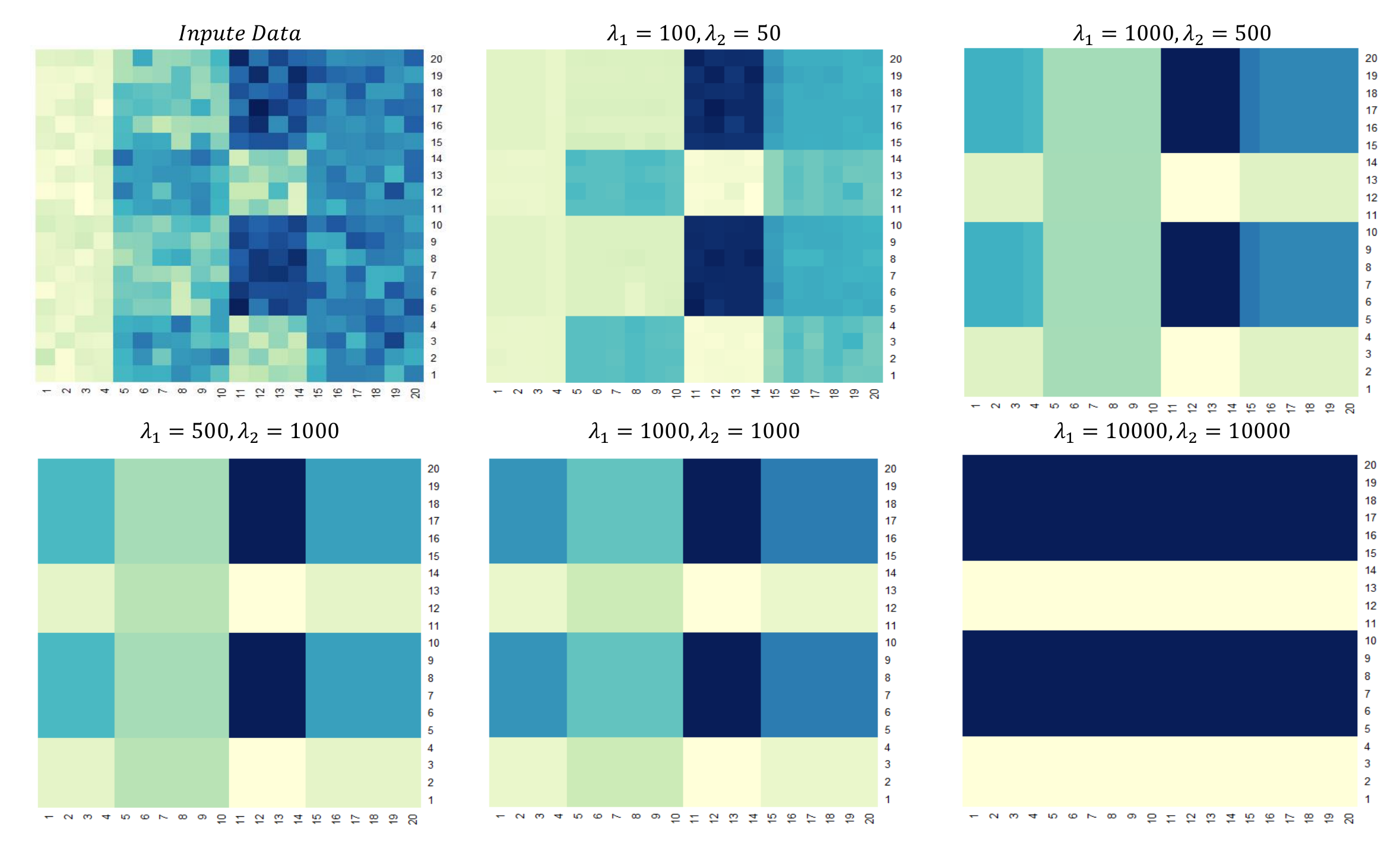}
	\caption{Results of SUBIC method implementation on the simulated data}
	\label{fig:sim}
\end{figure*}

 As depicted in Figure \ref{fig:sim}, tuning parameters provide a flexible mechanism to analyze data with both high and low variances. It is obvious that by increasing $\lambda_{1}$ and $\lambda_{2}$, rows and columns are unified to mean in each bicluster but when $\lambda_{1}$ and $\lambda_{2}$ get larger values such as 10000, bicluster patterns are ``smoothed out" and the number of biclusters reduces.

\begin{figure}[H]
	\centering
	\includegraphics[scale= 0.52]{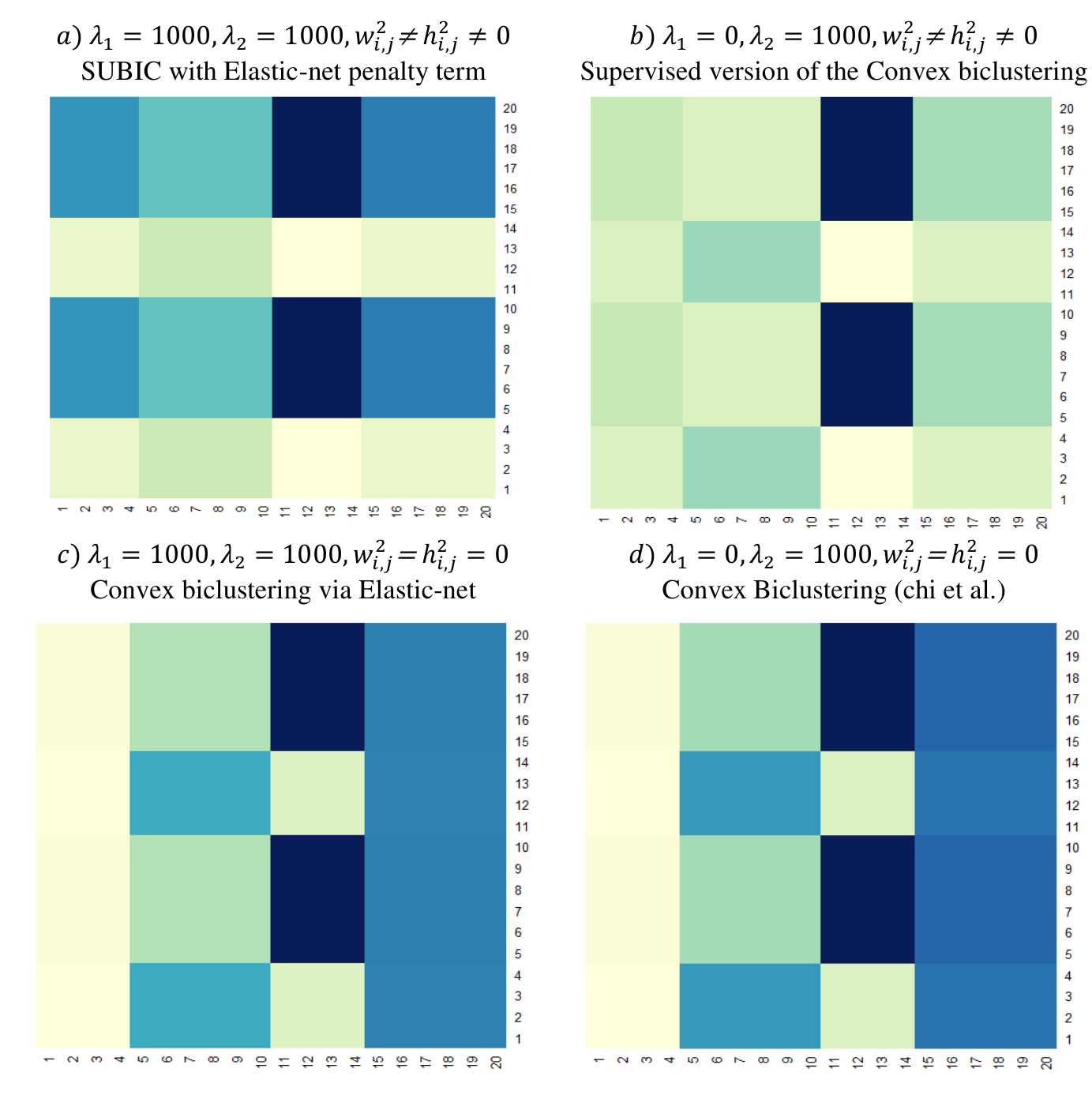} 
	\caption{Different scenarios which show the flexibility of SUBIC method}
	\label{fig:scenar}
\end{figure}

We consider different scenarios in Figure \ref{fig:scenar} to show that the flexibility and generalization of our method. Panel \textit{a} shows our supervised biclustering approach, SUBIC, with elastic-net penalty ($l_1$ and $l_2$) as the most general case. By zeroing out $\lambda_1$, the $l_2$ penalty (special case 1), SUBIC becomes the extended (supervised) version of the convex biclustering approach \cite{chi2016convex} (Panel \textit{b}). If we instead zero out the supervised weight components $w_{i,j}^2$ and $h_{i,j}^2$ (special case 2), SUBIC becomes extended unsupervised convex biclsutering with elastic-net penalty (Panel \textit{c}). Finally, if we zero out both the $l_2$ penalty and the supervised weight components $w_{i,j}^2$ and $h_{i,j}^2$ (special case 3), SUBIC becomes the {\it bona fide} convex biclustering method reported in \cite{chi2016convex}. 
Therefore, our SUBIC approach is sufficiently general and flexible that employs a target value to guide the subgroup detection by encouraging sparsity of the number of variable groups and variables within each group. Correspondingly, our SUBIC approach most accurately detect the biclusters given in the ground truth. Panel \textit{a} and \textit{b} in Figure \ref{fig:scenar} confirm that the impact of supervised weights (target value guidance) in identifying of true biclusters in comparison with convex biclustering approach \cite{chi2016convex} (Panels \textit{c} and \textit{d}). Also in both cases the elastic-net regularization appears more accurate in detecting true biclusters.     

We extend the above simulation idea to $80\times80$ matrix and consider different design (true biclusters) with two variance levels (low and high) for assessment the performance of our model. We use different tuning parameters in each design and evaluate SUBIC method in terms of rand index and adjusted rand index. The results of average RI and ARI over 10 replicates are displayed in Table \ref{table 4} and \ref{table 5} for low-variance and high-variance data respectively.

\begin{table}[H]
	\tiny
	\caption{RI and ARI for different designs with low noisy simulated data}
	\label{table 4}
		\centering
		
		\begin{tabular}{c c | p{0.2cm} p{0.2cm} | p{0.2cm} p{0.2cm}| p{0.2cm} p{0.2cm}}
			\hline
			Design	&\ $\sigma$	&\multicolumn{2}{l|}{$\lambda_{1}=\lambda_{2}=10^{2}$ }  
			&\multicolumn{2}{l|}{$\lambda_{1}=\lambda_{2}=10^{3}$ }  
			&\multicolumn{2}{l}{$\lambda_{1}=\lambda_{2}=10^{4}$ }  
			\\
			& &RI	&ARI &RI	&ARI &RI	&ARI\\
			\hline
			$2\times4$  &1.5	& 0.85	& 0.71	&0.99	& 0.96 & 0.79	& 0.65	\\
			$4\times4$  &1.5	& 0.79	& 0.62	&0.98	& 0.95 & 0.76	& 0.64	\\
			$4\times8$  &1.5	& 0.73	& 0.56	&0.98	& 0.97 & 0.68	& 0.59	\\
			$8\times8$  &1.5	& 0.82	& 0.69	&0.96	& 0.93 & 0.72	& 0.61	\\

			\hline
		\end{tabular}
	\hfill
\end{table}

\begin{table}[H]
	\tiny
	\caption{RI and ARI for different designs with high noisy simulated data}
	\label{table 5}
	\centering
		\begin{tabular}{c c | p{0.2cm} p{0.2cm} | p{0.2cm} p{0.2cm}| p{0.2cm} p{0.2cm}}
			\hline
			Design	&\ $\sigma$	&\multicolumn{2}{l|}{$\lambda_{1}=\lambda_{2}=10^{2}$ }  
			&\multicolumn{2}{l|}{$\lambda_{1}=\lambda_{2}=10^{3}$ }  
			&\multicolumn{2}{l}{$\lambda_{1}=\lambda_{2}=10^{4}$ }  
			\\
			& &RI	&ARI &RI	&ARI &RI	&ARI\\
			\hline
			$2\times4$  &3	& 0.65	& 0.53	& 0.90	& 0.88 & 0.99	& 0.96	\\
			$4\times4$  &3	& 0.68	& 0.58	& 0.85	& 0.81 & 0.99	& 0.97	\\
			$4\times8$  &3	& 0.59	& 0.49	& 0.93	& 0.90 & 0.98	& 0.93	\\
			$8\times8$  &3	& 0.55	& 0.43	& 0.87	& 0.82 & 0.99	& 0.95	\\
			
			\hline
		\end{tabular}
\end{table}

As shown above, the performance of SUBIC is fully tunable using the pair of tuning parameters in response to  data with different levels of variances. From Table \ref{table 4} and \ref{table 5}, it is clear that SUBIC's superior performance is very stable for both low and high variance data. In particular, the robust performance against high-variance data is achieved by setting larger values of tuning parameters. 

\section{Application in Personalized Medicine}
In this section we demonstrate how SUBIC method is capable of identifying patient subgroups with guidance of the target variable LVMI. We study the population of African-Americans with hypertension and poor blood pressure control who have high risk of cardiovascular disease. 

\begin{figure}[H]
	\centering
	\captionsetup{justification=centering,margin=.07cm}
	\includegraphics[scale= 0.059]{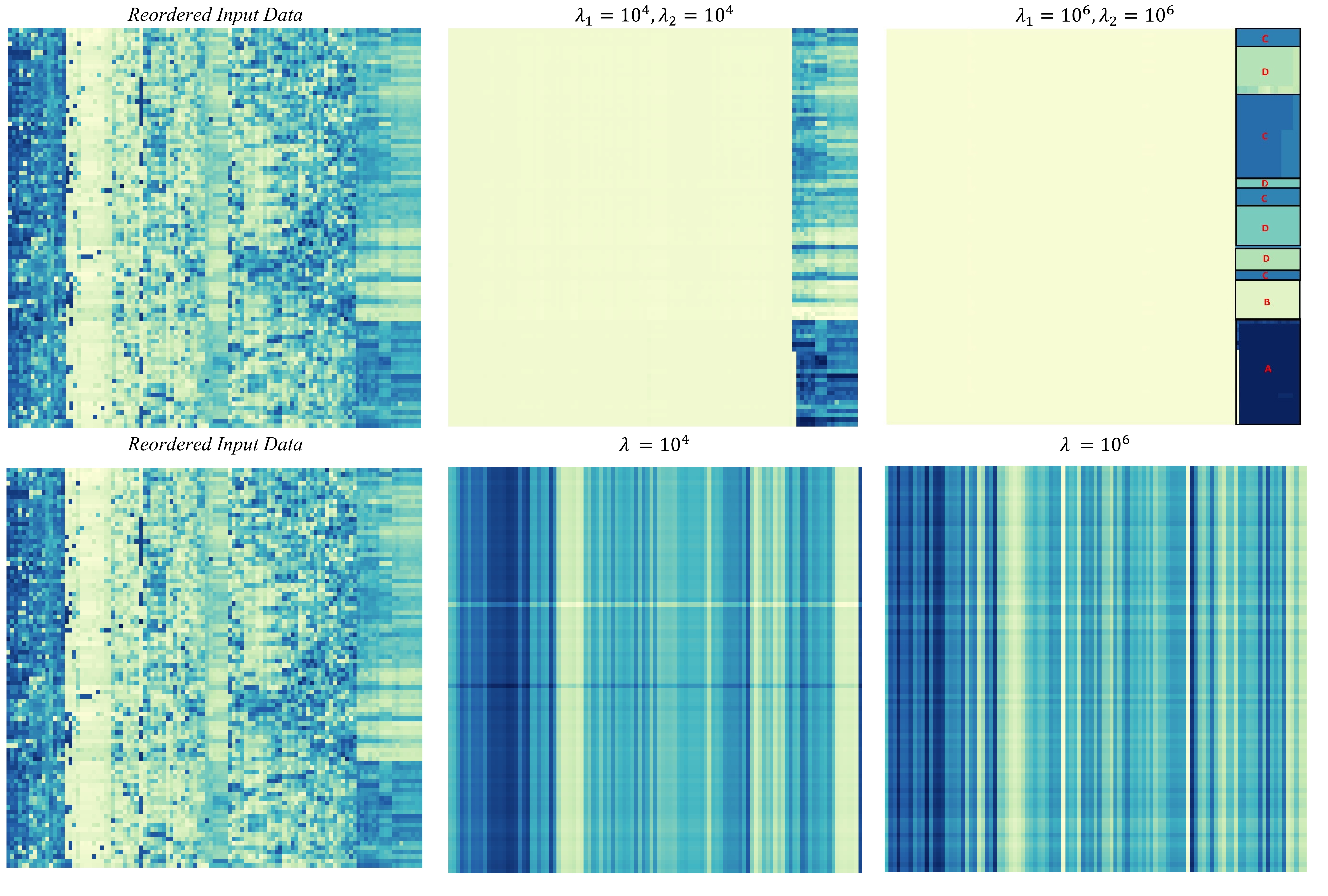} 
	\caption{Results of SUBIC implementation (top panel) and COBRA method (bottom panel) on the data related to African-American patients at- high risk of cardiovascular disease.}
	\label{figure4}
\end{figure}

Data are obtained from patients enrolled in the emergency department of Detroit Receiving Hospital. After preprocessing step, our data consists of 107 features including demographic characteristics, previous medical history, patient medical condition, laboratory test result, and CMR results related to 90 patients. To achieve a checkerboard pattern, we reorder rows and columns (original data) at first \cite{chi2016convex} using hierarchical clustering and then apply SUBIC method. The results are shown in the top panel of Figure \ref{figure4}. In addition, we implemented convex biclustering method (COBRA) developed by Chi et al. \cite{chi2016convex} using package ``cvxbiclustr" in R for comparing with our SUBIC method. Results obtain
ed using different tuning parameters ($\lambda$) are shown in the bottom panel of Figure \ref{figure4}.

In Figure \ref{figure4}, our SUBIC method detects 4 subgroups using 15 features for $\lambda_{1} = \lambda_{2}= 10^{4}$. These 15 features belong to 3 major groups of features including: 1) Waist Circumference Levels (mm); 2) Average Weight (kg) and 3) Calculated BMI. The statistics related to these risk factors based on 4 groups of patients is summarized in Table \ref{table 6}. It is worth mentioning that other potential risk factors such as ``Troponin Level" or ``Plasma Aldosterone" can be also significant but these three groups of features are sufficient to describe the disparity among patients based on guidance of the target variable LVMI. On the contrary, COBRA method fails to find any patient subgroups for this data set based on different tuning parameters.
\\

\begin{table}[H]
	\tiny 
	\centering
	\caption{Average of three disparity factors and LVMI (along with standard deviation) for subgroups detected by SUBIC}
	\label{table 6}
	\begin{tabular}{|c | c |c |c | c |c |c|}
		\hline
		Subgroup & size	& Waist Circumference (mm)  & Average Weight (kg) & Calculated BMI & LVMI \\ 
		\hline
		A  & 24	& 1248.8 (104.7)	& 125.1 (13.2 )& 41.6 (5.2)	& 85.7 (11.9) 	\\
		B  & 28	& 1092.6 (74.5)	& 99.7 (11.1)  	& 35.1 (3.8)	& 82.7 (13.7)		\\
		C  & 29	& 972.8 (89.6)	& 84.8 (10.2)	    & 30.1 (4.9)	& 80.9 (13.7)	\\ 
		D  & 9	& 813.3 (123.7)	& 64.4 (10.6)	    & 23.8 (4.3)	& 79.3 (11.8)	\\   
		
		\hline
		Total  & 90	& 1067.7 (163.5)	& 98.2 (22.2)	& 34.1 (7.2)	& 82.6 (12.9)	 	\\

		\hline
	\end{tabular}
\end{table}


\section {Discussion and Conclusion }
In this paper, we have developed a novel supervised subgroup detection method called SUBIC based on convex optimization. 
SUBIC is a predictive model that combines the strength of biclustering and tree-based methods. We introduced a new elastic-net penalty term in our model and defined two new weights in our objective function to enable the supervised training under the guidance of a clinically relevant target variable in detecting biclusters. We further presented a generalized additive model for predicting target variables for new patients. We evaluated our SUBIC approach using simulation studies and applied our approach to identify disparities among African-American patients who are at high risk of cardiovascular disease. Future directions include extending our SUBIC approach to predict categorical target variables, such as stages and subtypes of heart diseases.


	\bibliographystyle{plain}
	\bibliography{References}{}

\end{document}